\documentclass[11pt]{article}

\usepackage[preprint]{acl}

\usepackage{times}
\usepackage{latexsym}

\usepackage{booktabs} 
\usepackage{tcolorbox}
\usepackage{adjustbox}
\usepackage{amsmath}
\usepackage{amssymb}
\usepackage[T1]{fontenc}
\usepackage[utf8]{inputenc}

\usepackage{microtype}

\usepackage{inconsolata}

\usepackage{graphicx}

\title{SA-RAG: Structuring Noisy Conversations for\\ Retrieval-Augmented Generation}

\author{
  Kaiqiao Han$^{1,2}$, LuAn Tang$^{2}$, Renliang Sun$^{1,2}$, Peng Yuan$^{2}$, \\
  \textbf{Wei Cheng}$^{2}$, \textbf{Haoyu Wang}$^{2}$, \textbf{Wei Wang}$^{1}$, \textbf{Yizhou Sun}$^{1}$, \textbf{Haifeng Chen}$^{2}$ \\
  \normalfont $^1$UCLA \qquad $^2$NEC Labs \\
  \small
\begin{tabular}{c}
    \texttt{\{kqhan, sunrenliang, weiwang, yzsun\}@cs.ucla.edu} \\
    \texttt{\{ltang, pengyuan, weicheng, haoyuwang, Haifeng\}@nec-labs.com}
  \end{tabular}
}
\begin{document}
\maketitle
\begin{abstract}
% Large Language Models (LLMs) have been widely adopted in conversational applications. However, their reliance on parametric knowledge limits their reliability in real-world scenarios that require access to dynamic and domain-specific information. Retrieval-Augmented Generation (RAG) addresses this issue by incorporating external knowledge during generation, but existing text-based and graph-based RAG approaches often fail in practice due to poor-quality contexts with noisy and irrelevant information.
% In this work, we propose \emph{Structure-aware Retrieval Augmented Generation} (\textbf{SA-RAG}), focusing on tables as an intermediate structured representation and providing a compact and controllable interface that reduces noise while preserving essential information.
% We propose a quality-aware table metadata generation framework that explicitly models metadata normalization and metadata effectiveness, leading to improved metadata quality and enhanced downstream effectiveness.
% Moreover, we provide both training-free and training methods in table generation. The generation validation and direct preference optimization could improve the table quality and maintain the semantic and structure consistency in generation. Experiments on two noisy real-world datasets demonstrate that Structure RAG significantly outperforms RAG baselines in generation quality. Our code is publicly available at an anonymous repository~\footnote{\url{https://anonymous.4open.science/r/SA-RAG-E789}}.
Large Language Models (LLMs) have been widely adopted in conversational applications. However, their reliance on parametric knowledge limits reliability in real-world scenarios that require dynamic or domain-specific information. Retrieval-Augmented Generation (RAG) addresses this limitation by incorporating external knowledge during generation, but existing RAG methods often struggle with noisy or irrelevant contexts as they typically assume that high-quality structures are already available.
In this work, we propose \emph{Structure-aware Retrieval Augmented Generation} (\textbf{SA-RAG}), a principled structure induction and generation framework that constructs reliable intermediate structures directly from noisy conversational data. SA-RAG uses tables as compact and controllable structured representations. We introduce a quality-aware structure induction framework that jointly models semantic and structural consistency. Furthermore, we explore both training-free and training-based structure generation strategies to further improve structure quality while maintaining fidelity.
Experiments on two noisy real-world datasets show that SA-RAG significantly outperforms existing RAG baselines. Further analysis highlights that high-quality structural representations are crucial for robust RAG applications. Our code is publicly available at an anonymous repository~\footnote{\url{https://anonymous.4open.science/r/SA-RAG-E789}}.
\end{abstract}

\section{Introduction}

Large Language Models (LLMs) are widely used in conversational applications such as customer service, virtual assistants, and question answering systems~\cite{liang2025llm,zhang2020dialogptlargescalegenerativepretraining,adiwardana2020humanlikeopendomainchatbot,Rome_2024}. They can maintain conversational context and generate coherent responses, improving user experience across domains~\cite{yang2025ragvaengineeringretrievalaugmented,xie2026securingllmasaservicesmallbusinesses,reuters2025verizon}. However, parametric knowledge alone is often insufficient, as many real-world scenarios require external and domain-specific information~\cite{cuconasu2024power,laban2024summary,bui2025kgcqrleveragingstructuredrelation}. In conversational settings, knowledge-seeking queries are frequently mixed with noisy or irrelevant content, making reliable retrieval essential~\cite{hu-etal-2025-removal}. Retrieval-Augmented Generation (RAG) addresses this challenge by incorporating information retrieved from external knowledge sources during generation~\cite{lewis2021retrievalaugmentedgenerationknowledgeintensivenlp,10.5555/106765.106782}.
\begin{figure*}[!t]
    \centering
    \includegraphics[width=\textwidth, trim=20 570 20 20, clip]{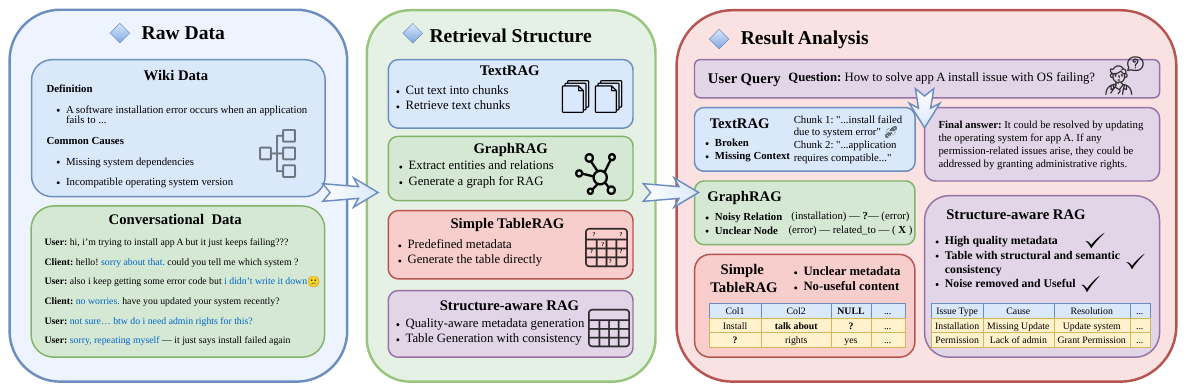}
    \caption{Comparison of SA-RAG with other methods. On clean and well-organized datasets (e.g., Wikipedia-style corpora), most existing RAG methods perform competitively. However, conversational data is inherently noisy, containing irrelevant turns and structural inconsistencies. In such settings, conventional approaches, such as text-based RAG, graph RAG, and simple table RAG, struggle to effectively retrieve and utilize relevant information. In contrast, our Structure-Aware RAG explicitly models and leverages underlying structural signals.}
    % \kq{big fig and text} 
    % \RL{Font sizes are inconsistent. Furthermore, I believe the issues with the other three existing RAG methods should be marked with an X icon or something similar to indicate errors. Currently, it doesn't sufficiently highlight the advantages of your approach.}}
    \label{fig:intro}
\end{figure*}

Text-based RAG~\cite{lewis2021retrievalaugmentedgenerationknowledgeintensivenlp,10.5555/106765.106782} retrieves passages from unstructured text, while graph-based RAG supports reasoning through entity--relation structures~\cite{moreira2024enhancing,edge2024local,laban2024summary}. However, real-world knowledge sources are often noisy and redundant, limiting both paradigms. Text-based RAG is sensitive to irrelevant context, whereas graph-based RAG~\cite{wang2025knowledgegraphaugmentedlarge,edge2024local} depends on accurate entity and relation extraction, which is difficult in noisy conversational settings. Existing information extraction approaches also rely on predefined schemas or rules and remain brittle under noisy dialogue data~\cite{mausam-etal-2012-open}. 
We argue that the core bottleneck of conversational RAG is not retrieval itself, but the lack of structurally consistent intermediate representations under noisy conversational evidence. Compared with free-form text and rigid graphs, tables provide an attribute-centric organization that aligns queries with semantically relevant fields while remaining robust to noisy, incomplete, or redundant content. 

% This observation motivates our central hypothesis: structurally consistent representations are critical for robust conversational RAG under noisy settings.

Motivated by these existing problems, we propose Structure-aware Retrieval-Augmented Generation (SA-RAG), a new RAG paradigm that uses tables as an intermediate structured representation between unstructured knowledge sources and LLM generation in noisy data. Tables provide an explicit and compact organization of information along meaningful dimensions, enabling effective noise reduction and more controllable grounding for generation. By serving as a structured interface, tables bridge the gap between raw text and fully symbolic representations, offering a favorable trade-off between expressiveness and robustness.

Introducing structured representations into conversational RAG raises a key challenge: how to induce reliable structures from noisy conversations. Unlike prior RAG settings that assume curated structures, real-world conversations contain fragmented information, redundant utterances, and heterogeneous expressions, making structure induction a critical bottleneck.
\textbf{(Q1) How can we induce high-quality structures for retrieval and generation?}
Existing methods either rely on manually designed schemas with limited generalization, or directly use LLMs for extraction, which often produce unstable outputs under noisy settings.
\textbf{(Q2) How can noisy evidence be normalized into a semantically consistent structure space?}
Conversational data often contains inconsistent attribute names, fragmented descriptions, and redundant context, leading to inconsistent representations of equivalent information.
\textbf{(Q3) How can induced structures maintain semantic and structural consistency?}
Since induced structures directly support retrieval, both schema misalignment and hallucinated values can significantly degrade downstream generation quality.

To address these challenges, we propose SA-RAG, a structure induction and generation framework for noisy RAG. Instead of assuming structured representations are readily available, SA-RAG explicitly constructs reliable intermediate structures from noisy conversational data. 
During structure induction, we introduce a quality-aware metadata construction framework that iteratively normalizes and refines candidate metadata, enabling semantically consistent alignment of noisy sources while preserving compact and informative structures. During structure generation, we explore both training-free and training-based strategies for metadata-aligned generation. The training-free approach enforces semantic and structural consistency, while the training-based approach adopts Direct Preference Optimization (DPO) to encourage high-quality structured outputs. Together, these designs improve structure quality and generation faithfulness in noisy real-world settings.

% Our contributions are summarized as follows:
% \begin{itemize}
%     \item We propose \textbf{Structure-aware RAG} (SA-RAG), a retrieval-augmented generation framework that formulates noisy conversational RAG as a \textbf{structure induction and generation problem}, constructing reliable intermediate structures directly from noisy multi-turn conversations for application.
    
%     \item We introduce a \textbf{quality-aware structure induction framework} and develop both training-free and training-based \textbf{structure generation strategies} to improve the quality and controllability of induced structures.
    
%     \item Experiments on multiple noisy conversational datasets show that SA-RAG \textbf{consistently outperforms} strong RAG baselines. We show that \textbf{structural quality is the dominant factor} for robust conversational RAG.
% \end{itemize}

Our contributions are summarized as follows:
\begin{itemize}
    \item We identify \textbf{structural quality under noisy evidence} as a key bottleneck of conversational RAG, and reformulate conversational retrieval-augmented generation as a \textbf{structure induction and generation problem} rather than a purely retrieval problem.
    
    \item We propose \textbf{Structure-aware RAG} (SA-RAG), a unified framework that induces structurally consistent intermediate representations directly from noisy multi-turn conversations through \textbf{quality-aware structure induction} and both training-free and training-based \textbf{structure generation strategies}.
    
    \item Through extensive experiments and robustness analysis on noisy conversational datasets, we demonstrate that SA-RAG consistently outperforms strong RAG baselines and show that \textbf{structural quality, rather than retrieval capacity alone, is the dominant factor for robust conversational RAG}.
\end{itemize}

\section{Related Work}

\subsection{Retrieval-Augmented Generation over Structured Data}

Recent RAG research extends retrieval from unstructured text to structured representations such as tables and knowledge graphs, improving evidence aggregation and multi-hop reasoning~\cite{cuconasu2024power,laban2024summary,bui2025kgcqrleveragingstructuredrelation}. Hybrid neural--symbolic approaches further enhance factual faithfulness and controllability by combining neural retrieval with structured operations. Most existing methods assume that structured representations are already available or manually curated~\cite{yin2020tabertpretrainingjointunderstanding,Herzig_2020}. In contrast, our work focuses on automatically inducing reliable structures directly from noisy conversational data.

\subsection{Content Quality Evaluation in RAG Systems}

Content quality is critical for RAG systems, as noisy or redundant evidence can degrade retrieval and increase hallucinations. Prior work mainly evaluates text-based RAG using query-aware metrics and improves reliability through reranking or filtering strategies~\cite{shi2024enhancing,cejas2025retrieval}. Unlike these approaches, which focus on unstructured documents, our method explicitly evaluates and regulates the quality of structured representations through metadata normalization and metadata effectiveness.
\section{Method}
\begin{figure*}[!t]
    \centering
    \includegraphics[width=\textwidth, trim=95 540 20 20, clip]{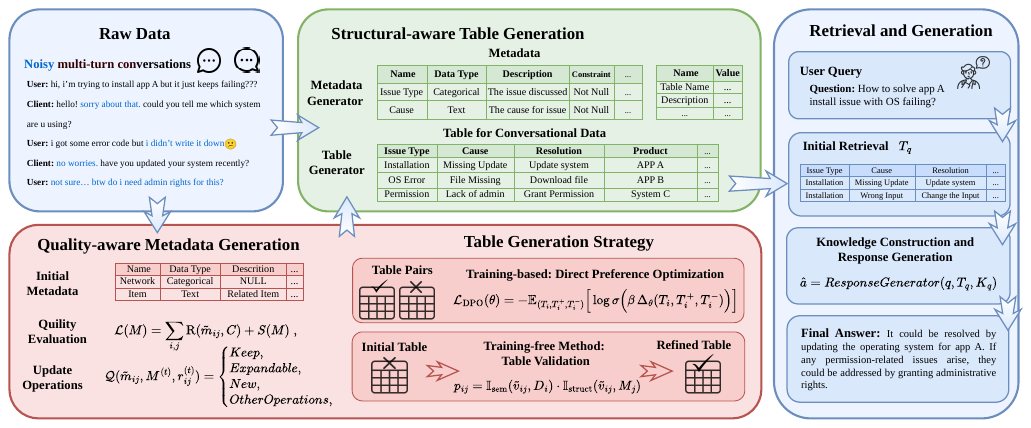}
    \caption{Model Overview. Given a user query and noisy multi-turn conversational history, Structure-Aware RAG first induces latent structured knowledge by generating metadata-aligned tables from raw conversations. The framework then performs joint retrieval over both raw conversations and their structured representations, enabling robust evidence selection under noise and redundancy. Finally, the generator produces a grounded response conditioned on the query and retrieved structured evidence.}
    \label{fig:model}
\end{figure*}
\subsection{Problem Definition}

We consider a conversational retrieval-augmented generation (RAG) setting
where the goal is to generate grounded responses to user queries under
noisy conversational data.

\paragraph{Input.}
The input consists of a user query $q$ with a historical conversational data
$\mathcal{D} = \{d_1, d_2, \ldots, d_N\}$, where $N$ is the number of the conversations.
Each conversation $d_i$ is a multi-turn conversation that may contain verbose descriptions,
irrelevant context, and redundant information with essential
knowledge scattered across turns.

\paragraph{Generation Objective.}
Given a query $q$, the system must identify relevant conversations  and information
to support the response generator in the response generation.
The final goal is to generate a response $a$ that is factually grounded
in the retrieved conversation evidence, rather than relying on parametric knowledge and hallucinated content.
Formally, we define the task as learning a mapping
$
f: (q, \mathcal{D}) \mapsto {a},
$
where $a$ is the answer supported by information contained in the historical conversations.

\subsection{Model Overview}

To address noise and redundancy problems in conversational RAG,
we propose \emph{Structure-aware Retrieval Augmented Generation} (\textbf{SA-RAG}),
a unified framework that explicitly models the latent structured knowledge
underlying noisy multi-turn conversations with tables.
Each conversation $d_i \in \mathcal{D}$
is associated with a structured representation $M_i$
that captures its essential information.
Formally,
\[
\mathcal{T} = \mathcal{G}_{\text{table}}(\mathcal{D}, M),
\qquad
\mathcal{T} = \{ T_1, T_2, \ldots, T_N \}.
\]
where each $T_i$ is a metadata-aligned set of attribute--value pairs for conversation $d_i$ and $M$ is the metadata.
These structures are not directly observable and must be inferred
from raw conversational text.

SA-RAG consists of three tightly coupled stages:
% SA-RAG consists of three tightly coupled stages for structure induction, structure alignment, and grounded response generation:

\begin{enumerate}
    \item \textbf{Quality-aware Structure Induction}, which constructs a normalized and high-quality metadata $M^{\star}$ from noisy conversations. 
    % This stage iteratively induces and refines metadata through normalization and quality-aware updates.

    \item \textbf{Structure Generation with Training-Free and Training-based Strategies}, which generates metadata-aligned tables $T_i$ for each conversation $d_i$.

    \item \textbf{Structure-aware Retrieval and Generation}, which identifies relevant structured evidence from tables and conversations, and produces the final grounded response $a$.
\end{enumerate}

\subsection{Quality-aware Structure Induction}

In structured databases, metadata describes the organization, schema, and properties of stored data, including table names, column names, data types, constraints, and relations between attributes.
% Within conversational RAG, metadata defines the latent structure space through which noisy conversational evidence is organized, aligned, and retrieved. Consequently, the quality of the induced metadata becomes a first-order factor that directly affects retrieval reliability and generation faithfulness.
However, inducing high-quality structures from noisy conversational data is fundamentally challenging. Unlike curated databases, real-world conversations contain fragmented descriptions, redundant utterances, and heterogeneous expressions~\cite{ramakrishnan2003database,eden2002metadata}.
% Even semantically identical concepts may appear with different granularities or surface forms across conversations. 
As a result, na\"ively aggregating metadata in structure induction often produces redundant, unstable, and low-utility structures.
More importantly, existing LLM-based extraction approaches optimize local extraction correctness while overlooking global structure quality.
% Without explicit structure-level normalization and quality control, metadata growth becomes uncontrolled, leading to semantic drift, duplicated or inconsistent attributes, and weak alignment between induced structures and user queries.

To address these challenges, we formulate metadata construction as a quality-aware structure induction problem. Rather than directly aggregating extracted attributes, SA-RAG explicitly models structure quality through normalization, semantic consistency, and downstream usefulness. The goal is to induce a compact yet expressive structure space that preserves essential conversational evidence while filtering noisy or redundant information in conversations.
% To address these issues,
% we explicitly introduce a quality-aware evaluation mechanism
% that jointly considers semantic coverage, structural coherence,
% allowing the metadata to evolve in a controlled and interpretable manner.
% This design enables the metadata to remain both expressive and compact,
% which is critical for robust table-based retrieval and generation.
Given a set of noisy multi-turn conversations
$\mathcal{D} = \{d_1, d_2, \ldots\}$,
our objective is to construct a consistent and high-quality metadata for table generation and downstream retrieval and generation tasks.
We explicitly model metadata consistency and metadata evolution
through an iterative, quality-aware process.

\paragraph{Initial Metadata Extraction.}

We start from an initial metadata $M_0$.
For each conversation $d_i$, an LLM extracts a set of candidate
metadata and corresponding information:
$
M_i = \{ (m_{ij}, b_{ij}) \mid j = 1, \ldots, |M_i| \},
$
where $m_{ij}$ denotes a candidate column name
and $b_{ij}$ its additional information.
e.g., type, semantic information.
Due to noisy phrasing and heterogeneous expressions, the extracted columns across conversations may be useless,
semantically overlapping or inconsistent, which restricts the table quality and the performance.

% Constructing structured table metadata directly from real-world noisy data
% is inherently challenging due to noisy language, incomplete description,
% and highly heterogeneous expressions.
% Unlike curated databases, noisy data often exhibits inconsistent attribute naming,
% and varying levels of granularity, even for semantically identical concepts.
% As a result, naïvely aggregating metadata extracted from individual data 
% tends to produce redundant and low-utility metadata,
% which severely degrades table generation quality and downstream
% retrieval-augmented generation performance.

\paragraph{Quality Evaluation and Update.}
To improve metadata consistency and quality, we normalize extracted column names
by mapping them to canonical forms following the practice of the database using LLMs.
Formally, we use a normalization function
$
\mathcal{N} : m_{ij} \mapsto \tilde{m}_{ij},
$,
which resolves synonymy, granularity mismatch, and formatting variations.
Applying $\mathcal{N}$ to all extracted columns yields
a normalized metadata set for each conversation:
$
\tilde{M}_i = \{ (\tilde{m}_{ij}, b_{ij}) \mid (m_{ij}, b_{ij}) \in C_i \}.
$
Then, we formulate metadata induction as an iterative optimization problem
that balances coverage and compactness.
Let $M^{(t)}$ denote the metadata at iteration $t$, and
$\tilde{m}_{ij}$ be a normalized column extracted from the $i$-th
conversation.
We define the following objective function:
$
\mathcal{L}(M)
=
\sum_{i,j}
\text{R}(\tilde{m}_{ij}, C)+S(M)
\;,
$
where $\text{R}(\cdot,\cdot)$ measures how well M could cover the useful information in $\tilde{m}_{ij}$, $S$ measures structural normalization quality as table metadata.

The optimization proceeds iteratively over $\mathcal{D}$. 
% \RL{Consider improvement rather than optimization? Optimization typically involves parameter adjustments.} 
At each step, a candidate normalized column is admitted only after passing quality validation, which ensures the update does not degrade the objective.
$
\Delta(\tilde{m}_{ij} \mid M^{(t)})
=
\mathcal{L}\!\left(M^{(t)} \oplus
 \{\tilde{m}_{ij}\}\right)
-
\mathcal{L}\!\left(M^{(t)}\right),
$
where $\oplus$ is one of the metadata operations, which are defined later.
At iteration $t$, for each $\tilde{m}_{ij}$, we compute its quality evaluation, e.g., answerability, relevance, and alignment 
with the current metadata:
$
r_{ij}^{(t)} = r(M^{i-1},\tilde{m}_{ij})
$
The benefit of incorporating $\tilde{m}_{ij}$ into the metadata is measured
by its gain. According to the quality of new metadata and the overall structure, we classify each column into one of the following metadata operations.
\begin{equation}
\mathcal{Q}(\tilde{m}_{ij}, M^{(t)},r_{ij}^{(t)})=
\begin{cases}
\textsc{Keep}, &
\\
\textsc{Expandable}, &
\\
\textsc{New}, &
\\
\textsc{Other OPerations
}, &
\end{cases}
\end{equation}
where $\mathcal{Q}$ is the metadata operations to update the metadata, we refer readers to the Appendix \ref{app.implement} for details and examples of the operations.
The metadata is updated by greedily incorporating columns that yield
positive improvements:
$
M^{(t+1)}
=
M^{(t)} \oplus
\big\{
\tilde{m}_{ij}
\;\big|\;
\Delta(\tilde{m}_{ij} \mid M^{(t)}) > 0
\}
$ to maintain the quality of the metadata. To balance reserving most information and metadata complexity, we set a maximum capacity for the metadata.
% \kq{ref}.
% This procedure monotonically increases $\mathcal{L}(M)$.

\paragraph{Iterative Construction.}
Processing all conversations sequentially induces
a sequence of metadata updates
$\{M^{(0)}, M^{(1)}, \ldots\}$,
with $M^{(0)} = M_0$.
After the processing, we will have the metadata
$M^{\star}$,
which represents a high-quality, normalized, and consistent
metadata for the conversation input.

% Our quality-aware construction ensures high-quality metadata by jointly enforcing query relevance and structural consistency during metadata evolution. Candidate columns are admitted only if they improve semantic coverage and answerability with respect to user queries, preventing the accumulation of irrelevant or noisy attributes. Meanwhile, normalization and structure-aware evaluation control schema growth and redundancy, yielding a compact, consistent metadata representation that is well aligned with downstream table construction and retrieval-augmented generation.
\subsection{Structure Generation Strategy with Semantic and Structural Consistency}

% Structured tables provide an explicit and machine-interpretable
% representation of knowledge,
% serving as a critical bridge between unstructured data
% and downstream retrieval and generation modules.
% Compared with free-form text,
% Well-constructed tables enable precise knowledge extraction and retrieval,
% which are essential for robust performance.
% However, generating high-quality tables requires faithful grounding of values while strictly preserving semantic and structural consistency with the global metadata.

% Table generation errors typically arise from two complementary failure modes:
% \emph{semantic inconsistency}, where values are hallucinated,
% inferred beyond the conversation, or weakly supported by evidence;
% and \emph{structural inconsistency}, where values violate column semantics,
% type constraints, or schema alignment.
% Either form of inconsistency can significantly degrade downstream performance,
% leading to unreliable retrieval
% and erroneous generation results~\cite{liu-etal-2025-quasar,ji2025targetbenchmarkingtableretrieval}.
% Therefore, effective table construction must jointly optimize
% semantic faithfulness to the source conversation
% and structural compatibility with the finalized metadata.
% This dual requirement motivates our table generation strategy,
% which explicitly enforces semantic validation and structural constraints,
% and further refines generation behavior through preference-based optimization.

Structured tables provide an interpretable representation bridging unstructured conversations and downstream retrieval and generation. High-quality structure generation requires both semantic faithfulness and structural consistency with metadata. Errors arise from semantic inconsistencies (hallucinated or weakly supported values) and structural violations (type or schema misalignment), which degrade retrieval and generation~\cite{liu-etal-2025-quasar,ji2025targetbenchmarkingtableretrieval}. Our structure generation strategy addresses these issues via semantic and structural validation and preference-based optimization.

Given the finalized high-quality metadata
$M^{\star} = \{m_1, m_2, \ldots, m_{|M|}\}$,
our goal is to generate a structured table for each conversation $D_i$
that is both semantically faithful to the original conversation
and structurally consistent with the metadata.
Each table row $T_i$ corresponds to a single conversation
and is represented as a set of attribute--value pairs:
$
T_i = \{ (m_j, v_{ij}) \mid m_j \in M^{\star} \}.
$
Considering the computational cost,
we propose both training-free and training-based generation strategies to improve the semantic and structural consistency.

\paragraph{Training-free Structure Generation.}
In the training-free setting, table values are generated directly by an LLM.
For each conversation $D_i$ and each metadata column $M_j \in M^{\star}$,
the value generation process is defined as
$
\tilde{v}_{ij} = \mathcal{G}_{\text{table}}(d_i, c_j),
$
yielding a candidate table
$\tilde{M}_i = \{(m_j, \tilde{v}_{ij})\}$.
% To ensure quality, we apply two validation constraints.
% First, \emph{semantic validation} verifies whether the generated value
% is supported by the original conversation:
% \begin{equation}
% \mathbb{I}_{\text{sem}}(\tilde{v}_{ij}, D_i) =
% \begin{cases}
% 1, & \tilde{v}_{ij} \ \text{is entailed by } D_i, \\
% 0, & \text{otherwise}.
% \end{cases}
% \end{equation}
% Second, \emph{structural validation} enforces adherence to the metadata
% in terms of format, type, and column alignment:
% \begin{equation}
% \mathbb{I}_{\text{struct}}(\tilde{v}_{ij}, m_j) =
% \begin{cases}
% 1, & \tilde{v}_{ij} \ \text{conforms to metadata } m_j, \\
% 0, & \text{otherwise}.
% \end{cases}
% \end{equation}
% Only values that satisfy both constraints are retained. The value will be dropped if it does not satisfy the constraints:
% \begin{equation}
% p_{ij}
% =
% \mathbb{I}_{\text{sem}}(\tilde{v}_{ij}, D_i)
% \cdot
% \mathbb{I}_{\text{struct}}(\tilde{v}_{ij}, M_j),
% \qquad
% v_{ij}
% =
% p_{ij} \odot \tilde{v}_{ij}.
% \end{equation}
% Values with $p_{ij}=0$ are treated as missing entries (\textsc{NULL}) rather than numerical zeros.
% The resulting validated table is denoted as $T_i$.
To ensure table quality, we apply \emph{semantic} and \emph{structural} validation. A generated value $\tilde{v}_{ij}$ is retained only if it is supported by the conversation $D_i$ and conforms to metadata $m_j$:
\[
p_{ij} = \mathbb{I}_{\text{sem}}(\tilde{v}_{ij}, D_i) \cdot \mathbb{I}_{\text{struct}}(\tilde{v}_{ij}, m_j), \quad
v_{ij} = p_{ij} \odot \tilde{v}_{ij}.
\]
Values failing either check ($p_{ij}=0$) are treated as missing (\textsc{NULL}). After the process, we have the resulting validated table $T_i$.

\paragraph{Training-based Structure Generation.}
While training-free generation enforces constraints post-hoc, it cannot proactively shape outputs, leading to low-quality or borderline tables in ambiguous cases. To address this, we adopt a training-based approach using Direct Preference Optimization (DPO)~\cite{rafailov2024directpreferenceoptimizationlanguage}, which aligns table generation with both semantic faithfulness and metadata consistency. For each conversation $d_i$, we construct a preference pair $(M_i^{+}, M_i^{-})$, where $M_i^{+}$ is a high-quality human-annotated table and $M_i^{-}$ is generated via controlled perturbations introducing semantic or structural noise. DPO contrasts these pairs to provide a fine-grained learning signal, enabling globally coherent and robust table generation (details in Appendix~\ref{app.implement}).
% \RL{Lacks specific implementation details. Lack of quality control description.}

Given a policy model $\pi_\theta$ and a frozen reference model
$\pi_{\text{ref}}$, the DPO objective is defined as
\begin{equation}
\scalebox{0.85}{$
\mathcal{L}_{\mathrm{DPO}}(\theta)
=
- \mathbb{E}_{(T_i, T_i^{+}, T_i^{-})}
\Big[
\log \sigma \big(
\beta \, \Delta_\theta(T_i, T_i^{+}, T_i^{-})
\big)
\Big],
$}
\end{equation}
\begin{equation}
\scalebox{0.85}{$
\Delta_\theta(T_i, T_i^{+}, T_i^{-})
=
\log \frac{\pi_\theta(T_i^{+} \mid T_i, M^{\star})}
{\pi_{\mathrm{ref}}(T_i^{+} \mid T_i, M^{\star})}
-
\log \frac{\pi_\theta(T_i^{-} \mid T_i, M^{\star})}
{\pi_{\mathrm{ref}}(T_i^{-} \mid T_i, M^{\star})}.
$}
\end{equation}

where $\beta$ is a temperature hyperparameter.
Optimizing this objective encourages the model to generate tables
that are preferred in terms of both semantic fidelity
to the conversation and structural consistency with $M^{\star}$.
After training, the model is used to generate the final table collection
$
\mathcal{T} = \{ T_1, T_2, \ldots \},
$
where each $T_i$ is a high-quality structured row of the table
corresponding to conversation $D_i$.

\subsection{Structure-aware Retrieval and Generation}

Given the structured tables $\mathcal{T}$ and a query $q$, we perform dense retrieval by embedding tables and query into a shared space and selecting top candidates. Information extraction proceeds iteratively: at step $t$, active values $E^{(t)}$ and relations $R^{(t)}$ are expanded via table lookups, forming the accumulated information $I^{(t)}$. The process terminates when sufficient evidence is collected, yielding structured triples $K_q = \{(e,r,i)\}$ grounded in table entries. The final response is generated by conditioning the LLM on $q$ and $K_q$, ensuring outputs are grounded in retrieved and verified structured evidence:
\[
\hat{a} = \text{Response Generator}(q, \mathcal{T}_q, K_q).
\]

\section{Experiments}
\begin{table*}[ht]
\centering

\begin{adjustbox}{max width=\textwidth}
\begin{tabular}{lcccccccccccc}
\toprule
\textbf{Model} 
& \multicolumn{6}{c}{\textbf{MSDialog}} 
& \multicolumn{6}{c}{\textbf{CSDialog}} \\
\cmidrule(lr){2-7} \cmidrule(lr){8-13}
& \textbf{Recall} & \textbf{↑\%} & \textbf{MRR} & \textbf{↑\%} & \textbf{Correctness} & \textbf{↑\%} 
& \textbf{Recall} & \textbf{↑\%} & \textbf{MRR} & \textbf{↑\%} & \textbf{Correctness} & \textbf{↑\%} \\
\midrule
TextRAG-freq 
& 0.52 & -- & 0.40 & -- & 0.51 & -- 
& 0.23 & -- & 0.14 & -- & 0.22 & -- \\

TextRAG-emb 
& 0.70 & 34.6\% & 0.50 & 25.0\% & 0.70 & 37.3\%
& 0.14 & -39.1\% & 0.08 & -42.9\% & 0.14 & -36.4\% \\

GraphRAG-KG 
& 0.18 & -65.4\% & 0.15 & -62.5\% & 0.17 & -66.7\%
& 0.04 & -82.6\% & 0.02 & -85.7\% & 0.02 & -90.9\% \\

GraphRAG-rel
& 0.14 & -73.1\% & 0.09 & -77.5\% & 0.13 & -74.5\%
& 0.02 & -91.3\% & 0.01 & -92.9\% & 0.02 & -90.9\% \\

IE-rel
& 0.32 & -38.5\% & 0.22 & -45.0\% & 0.30 & -41.2\%
& 0.24 & 4.3\% & 0.16 & 14.3\% & 0.22 & 0.0\% \\

IE-rule
& 0.12 & -76.9\% & 0.09 & -77.5\% & 0.12 & -76.5\%
& 0.32 & 39.1\% & 0.19 & 35.7\% & 0.29 & 31.8\% \\

TableRAG-pre
& 0.18 & -65.4\% & 0.15 & -62.5\% & 0.16 & -68.6\%
& 0.26 & 13.0\% & 0.18 & 28.6\% & 0.25 & 13.6\% \\

TableRAG-dir
& 0.42 & -19.2\% & 0.27 & -32.5\% & 0.41 & -19.6\%
& 0.31 & 34.8\% & 0.25 & 78.6\% & 0.31 & 36.4\% \\

StructRAG
& 0.60 & 15.4\% & 0.36 & -10.0\% & 0.56 & 9.8\%
& 0.35 & 52.2\% & 0.20 & 42.9\% & 0.34 & 54.5\% \\

SA-RAG-qwen 
& 0.72 & 38.5\% & 0.67 & 67.5\% & 0.70 & 37.3\%
& 0.41 & 78.3\% & 0.28 & 100.0\% & 0.40 & 81.8\% \\

SA-RAG 
& \textbf{0.94} & \textbf{80.8\%} & \textbf{0.82} & \textbf{105.0\%} & \textbf{0.92} & \textbf{80.4\%} 
& \textbf{0.53} & \textbf{130.4\%} & \textbf{0.42} & \textbf{200.0\%} & \textbf{0.52} & \textbf{136.4\%} \\
\bottomrule
\end{tabular}
\end{adjustbox}
\caption{Comparison of different models  across three metrics with improvements (relative to Text-based RAG). We use Recall, MRR, and Correctness for evaluation, and all three metrics are the higher, the better.}
\label{main}
\end{table*}

\subsection{Experiment Setup}
We describe the experimental setup below; additional details are provided in Appendix~\ref{app.implement} and~\ref{app.data}.

\paragraph{Application Scenario and Datasets}
We study information-seeking conversational agents, where users describe problems in natural conversation, and systems must retrieve historical conversations to generate grounded responses. Such scenarios commonly arise in technical support and customer service. 
We evaluate on two datasets: \textbf{MSDialog} and the \textbf{Customer Service Dialogue Dataset}~\cite{InforSeek_User_Intent,InforSeek_Response_Ranking,InforSeek_User_Intent_Pred}. 

% MSDialog is a public corpus of information-seeking conversations, while the Customer Service dataset contains professional customer–service interactions where users report issues and agents provide solutions.

\paragraph{Baselines}
We compare against representative RAG baselines across four categories.
\textbf{Text-based RAG} retrieves unstructured documents, including Text-freq (term-frequency retrieval) and TextRAG-emb (dense embedding retrieval)~\cite{lewis2021retrievalaugmentedgenerationknowledgeintensivenlp,10.5555/106765.106782}. 
\textbf{Graph-based RAG} retrieves knowledge through graph structures, including GraphRAG-KG and GraphRAG-Rel~\cite{wang2025knowledgegraphaugmentedlarge,edge2024local}. 
\textbf{Information extraction (IE)} methods construct structured knowledge from text, including IE-rel and IE-rule~\cite{yates-etal-2007-textrunner,stanovsky-etal-2018-supervised}. 
\textbf{Table-based RAG} operates on tabular representations, including TableRAG-pre and TableRAG-dir~\cite{Herzig_2020,liu2022tapextablepretraininglearning}. \textbf{StructRAG}  identifies the optimal structure type for retrieval and generation to improve answer accuracy~\cite{li2024structragboostingknowledgeintensive}.

\paragraph{Implementation and Evaluation}
We use LLaMA-70B (Llama-3.3-70B-Instruct) and Qwen-1.5B (Qwen2-1.5B) as base models. For LLaMA-70B we apply training-free table generation, while Qwen-1.5B is trained with the proposed table generation objective. GPT-4o is employed as an \emph{LLM-as-a-judge}, with full human verification for the main table and 20\% human verification for the other results. We refer readers to Appendix~\ref{agreement} for a detailed analysis of using LLMs as judges.
We evaluate with \textbf{Recall} and \textbf{MRR} to measure retrieval effectiveness~\cite{lewis2021retrievalaugmentedgenerationknowledgeintensivenlp,karpukhin-etal-2020-dense}, and \textbf{Result Correctness} to assess answer factuality with respect to the retrieved evidence. The latter follows an LLM-as-a-judge protocol with manual validation~\cite{liu2025judgejudgeimprovingevaluation,ho2025llmasajudgereassessingperformancellms}.

\subsection{Main Results}

We evaluate our method on two noisy conversational datasets, with results shown in Table~\ref{main}. Across both MSDialog and CSDialogue, SA-RAG consistently outperforms all baselines in Recall, MRR, and correctness. The improvements are particularly significant under noisy conversational settings, supporting our central claim that \textbf{structure is a critical factor for robust conversational RAG}. 
Although CSDialogue contains substantially longer and more complex interactions, SA-RAG remains robust and achieves consistent gains even in training-free settings, while training-based structure generation further improves smaller models under efficiency constraints. Compared with text-based RAG, which is highly sensitive to noisy and weakly relevant evidence, SA-RAG constructs compact and structurally consistent intermediate representations that suppress irrelevant conversational content while preserving retrieval-critical attributes. Graph-based RAG methods are further limited by incomplete schemas and unstable relation extraction in conversational settings with implicit or fragmented evidence, while direct table generation often introduces hallucinated or structurally inconsistent values. 
In contrast, SA-RAG explicitly models structure induction and generation, enabling more reliable retrieval grounding and downstream response generation. Unlike StructRAG, which assumes predefined structures and focuses primarily on retrieval routing, SA-RAG addresses the more fundamental challenge of constructing high-quality structures directly from noisy conversations. The consistent improvements across all metrics suggest that conversational RAG performance depends not only on the structure used, but more fundamentally on the \textbf{quality of the structures}.

\subsection{In-depth Analysis about the Structure Quality Results}

\begin{figure}[!t]
    \centering
    % 将 width 改为 0.5\textwidth，即原来的一半
    \includegraphics[width=0.5\textwidth, trim=0 0 0 0, clip]{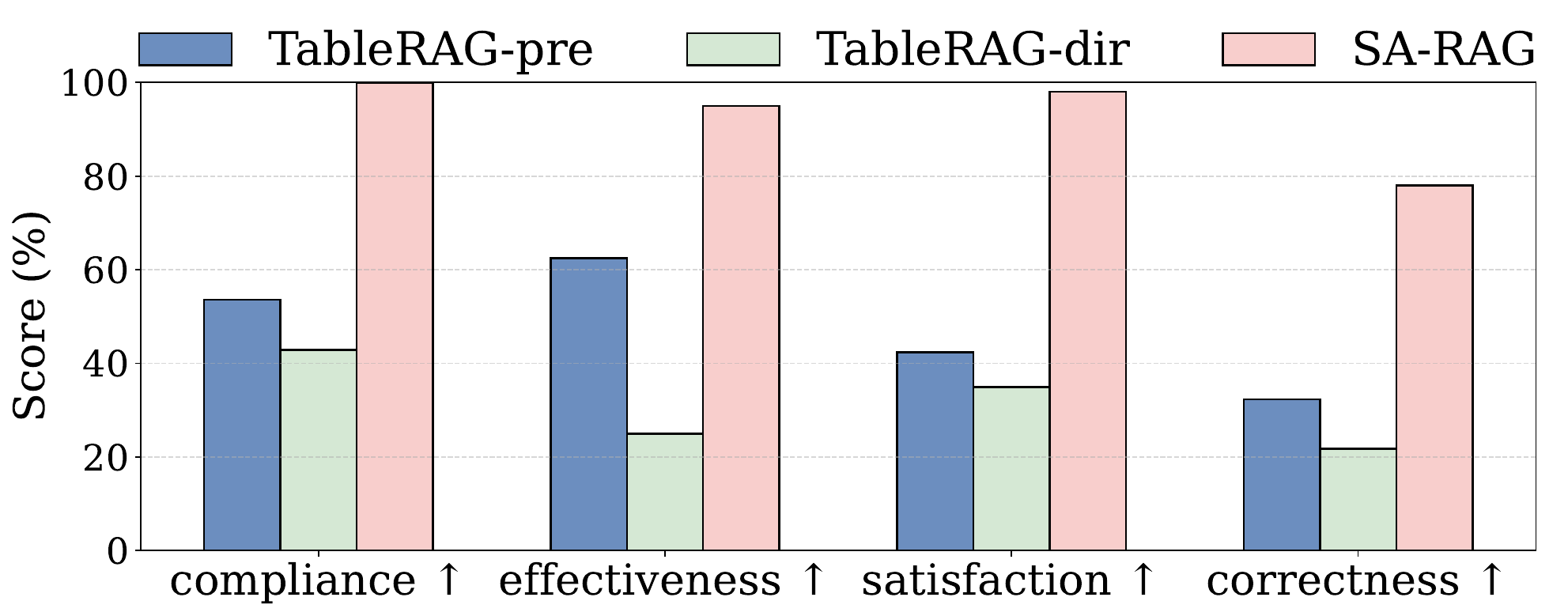}
    \caption{Comparison of Structure Quality Metrics Across Different Structure Construction Methods. For the four metrics, the higher, the better. }
     \vspace{-1.5em} % 或 -1em 视情况调
    % \RL{The diagram is not visually appealing. A legend on the right could be used to show what each color represents.} \RL{The title can be improved to `Comparison of table quality metrics across table construction methods. Each subfigure reports one metric and higher scores indicate better performance.'}}
    \label{fig:table}
\end{figure}

To better understand the role of \textbf{structural quality} in conversational RAG, we evaluate the generated tables using four complementary metrics: (1) \textbf{structural compliance}, (2) \textbf{metadata effectiveness}, (3) \textbf{constraint satisfaction}, and (4) \textbf{semantic correctness}. Detailed definitions and calculation methods are provided in Appendix~\ref{app.implement}. Figures~\ref{fig:table} summarize the results across all metrics. 
Overall, SA-RAG consistently outperforms both baselines, achieving stronger adherence to metadata constraints, more effective retrieval-oriented structures, fewer invalid or inconsistent entries, and higher semantic fidelity to the original conversations. More importantly, the results reveal a strong correlation between \textbf{structural consistency} and downstream conversational RAG performance. These findings support our central claim that \textbf{structural quality, rather than retrieval capacity alone, is a dominant factor for robust conversational RAG under noisy settings}.

\subsection{Ablation Study}

To assess each component's contribution, we conduct an ablation study with three SA-RAG variants on the same datasets and metrics. 
\textbf{SA-RAG w/o m} removes metadata construction, generating tables without normalization or effectiveness modeling. 
\textbf{SA-RAG w/o v} disables table validation, using all generated tables regardless of quality. 
\textbf{SA-RAG w/o d} removes DPO fine-tuning, relying on zero-shot generation. 
As shown in Table~\ref{tab:ablation}, all removals degrade performance. Omitting metadata (w/o m) most strongly reduces Recall and Eval, highlighting its role in structural guidance. Disabling validation (w/o v) lowers quality by allowing semantically inconsistent tables. Removing DPO (w/o d) causes the largest overall drop, especially for smaller models, indicating its importance in aligning table generation with downstream objectives.
Overall, the ablation study validates \textbf{effectiveness of modules in improving the structure quality}, and \textbf{structural quality} is important for achieving robust conversational RAG performance.
\begin{table}[t]

\centering
\begin{adjustbox}{max width=0.48\textwidth}
\begin{tabular}{lcc|cc|cc}
\toprule
Method
& \multicolumn{2}{c|}{Recall $\uparrow$}
& \multicolumn{2}{c|}{MRR $\uparrow$}
& \multicolumn{2}{c}{Correctness $\uparrow$} \\
 & Value & ↑\% & Value & ↑\% & Value & ↑\% \\
\midrule
SA-RAG (Llama)
& 0.94 & -
& 0.82 & -
& 0.92 & - \\

SA-RAG (Gemini)
& 0.92 & -2\%
& 0.81 & -1\%
& 0.90 & -2\% \\

SA-RAG (Qwen-1.5B)
& 0.72 & -23\%
& 0.67 & -18\%
& 0.70 & -24\% \\

SA-RAG (Qwen-8B)
& 0.89 & -5\%
& 0.81 & -1\%
& 0.88 & -4\% \\

SA-RAG w/o m
& 0.45 & -52\%
& 0.46 & -44\%
& 0.42 & -54\% \\

SA-RAG w/o v
& 0.87 & -7\%
& 0.72 & -12\%
& 0.71 & -23\% \\

SA-RAG w/o d
& 0.21 & -78\%
& 0.17 & -79\%
& 0.12 & -87\% \\
\bottomrule
\end{tabular}
\end{adjustbox}

\caption{Ablation study of SA-RAG and comparison of different models. We report performance and relative change compared to the SA-RAG (Llama).}
\label{tab:ablation}
\vspace{-10pt}
\end{table}

\subsection{ Generalization and Structure Robustness}

To further study the robustness of SA-RAG under different domain distributions, noise conditions, models, and other conditions, we conduct additional analyses from three perspectives: (1) out-of-domain generalization, (2) noise propagation robustness, and (3) training scalability and different LLM backbones. Additional analysis on (4) table value granularity is provided in Appendix~\ref{app:case_study}.
\paragraph{Out-of-Domain Generalization}
\begin{table}[!t]
\centering
\small
\setlength{\tabcolsep}{4pt}
\begin{tabular}{lccccc}
\toprule
Method & Recall$\uparrow$ & MRR$\uparrow$ & Corr.$\uparrow$ & ND$\downarrow$ & Gain$\uparrow$ \\
\midrule
TextRAG-freq & 0.00 & 0.00 & 0.02 & 0.98 & 0.00 \\
GraphRAG-KG & 0.20 & 0.13 & 0.18 & 0.79 & 0.16 \\
TableRAG-dir & 0.45 & 0.28 & 0.41 & 0.52 & 0.39 \\
SA-RAG & \textbf{0.60} & \textbf{0.52} & \textbf{0.57} & \textbf{0.34} & \textbf{0.55} \\
\bottomrule
\end{tabular}
\caption{
Evaluation results on the out-of-domain dataset.
}
\label{tab:amazon_results}
\end{table}
% To evaluate whether SA-RAG generalizes beyond conversational support scenarios, we further conduct experiments on the Amazon Reviews dataset, which contains substantially noisier and less structured user-generated content. Compared with task-oriented dialogues, review data exhibits high emotional density, sparse factual grounding, and highly heterogeneous expressions, making structure induction and retrieval significantly more challenging.
% Following the same retrieval and generation pipeline as in the main experiments, we construct a retrieval corpus consisting of 1,000 reviews and evaluate on 50 query instances. Results are shown in Table~\ref{tab:amazon_results}.

% SA-RAG consistently outperforms all baselines across retrieval and generation metrics. The results suggest that the proposed structure-aware framework is not limited to conversational support datasets, but generalizes to noisier open-domain scenarios. We attribute this robustness to the fact that SA-RAG models latent structural consistency rather than relying on domain-specific surface patterns.

To evaluate whether SA-RAG generalizes beyond conversational support scenarios, we further conduct experiments on the Amazon Reviews dataset, which contains substantially noisier and less structured user-generated content. Please refer to Appendix~\ref{ood} for details. Results in Table~\ref{tab:amazon_results} show that SA-RAG consistently outperforms existing baselines across different metrics, suggesting that SA-RAG can transfer to a non-conversation,
user-generated review setting. We attribute this robustness to the fact that \textbf{ SA-RAG models latent structural consistency} rather than relying on domain surface patterns.

\paragraph{Noise Propagation Analysis}
\begin{figure}[!t]
    \centering
    % 将 width 改为 0.5\textwidth，即原来的一半
    \includegraphics[width=0.5\textwidth, trim=0 0 0 0, clip]{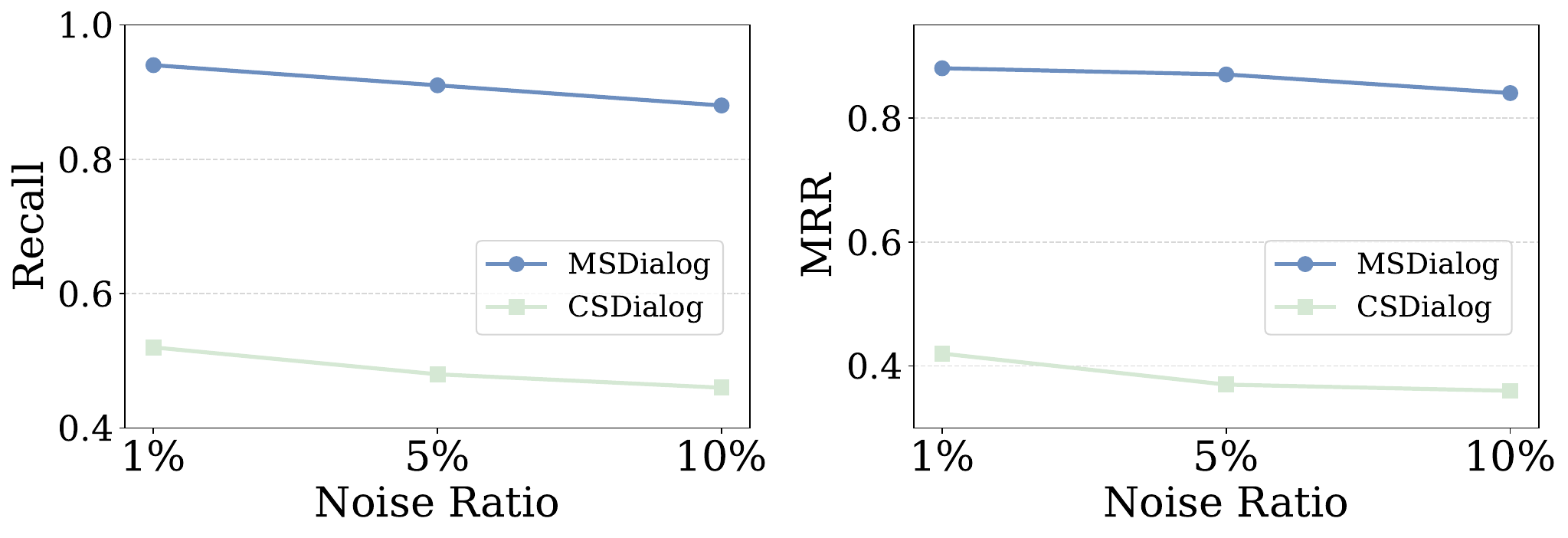}
    \caption{Noise propagation analysis under different noise injection ratios. }
     \vspace{-1em} % 或 -1em 视情况调
    % \RL{The diagram is not visually appealing. A legend on the right could be used to show what each color represents.} \RL{The title can be improved to `Comparison of table quality metrics across table construction methods. Each subfigure reports one metric and higher scores indicate better performance.'}}
    \label{fig:noise}
\end{figure}

Our method is explicitly designed to handle noise in conversational data by improving both semantic and structural consistency. To further analyze the influence of different noise levels, we conduct experiments by injecting noise into the generated tables. We mask different ratios of values in the table to add the noise and keep the other setup the same as the original experiment. The results show that increased noise leads to clear performance degradation. This finding further validates our motivation: The quality of structured representations is critical for retrieval performance. Without explicit structure quality control, noise indeed propagates and degrades retrieval. It also confirms that \textbf{ our improvements stem from noise reduction and structure refinement}, rather than relying solely on model capacity.

\paragraph{Training Scalability and Different LLM Backbones}

To further evaluate the scalability and generalization ability of SA-RAG, we conduct additional experiments using both larger models and alternative LLM backbones. Specifically, we scale our structure-aware DPO training pipeline from Qwen-1.5B to Qwen3-8B without introducing additional task-specific modifications, while keeping all other experimental settings unchanged. We also evaluate SA-RAG using Gemini (gemini-3-pro-preview) as an alternative backbone model.
As shown in Table~\ref{tab:ablation}, SA-RAG scales effectively with larger models, while smaller models remain competitive after structure-aware fine-tuning, demonstrating strong efficiency and practical applicability. Moreover, the performance remains stable across different backbone families, indicating that \textbf{the improvements primarily arise from quality-aware structure induction and generation framework rather than a particular LLM backbone.}

\subsection{Scalability and Adaptability}
Our framework efficiently supports large-scale conversation corpora with metadata optimization complexity of \(O(N)\), where \(N\) is the number of sampled conversations. Since metadata captures invariant domain-level structures, it can be induced from a representative subset and reused across the full corpus, substantially reducing practical computational cost.
The framework further supports distributed optimization on partitioned subsets, followed by centralized normalization and validation. For newly arriving dialogue turns, incremental updates are performed through localized table population without reprocessing the entire corpus. Please refer to Appendix~\ref{app:c} for more details.
Overall, by decoupling structure learning from online retrieval, SA-RAG achieves both scalability and adaptability in dynamic environments.
% \subsection{Case Study}

% % To illustrate the framework, the model first extracts high-quality metadata from the conversation and generates structured tables aligned with this metadata in an offline process. During online inference, it retrieves relevant table rows and conversation snippets conditioned on the user query, and produces a grounded response. For example, when asking about resolving an installation error, the system identifies the software, operating system, error code, and suggested fixes from the structured table, and generates a factual answer grounded in both the table and the conversation context. A detailed procedure case is provided in Appendix~\ref{app:case_study}.
% To illustrate the framework, SA-RAG first induces high-quality structures from noisy conversations through offline metadata and table construction. During inference, the model retrieves relevant structured evidence conditioned on the user query and generates a grounded response. For example, for an installation error query, SA-RAG identifies relevant software issues, operating systems, and suggested fixes from the induced tables to produce a factual response. A detailed case study is provided in Appendix~\ref{app:case_study}.

\section{Conclusion}

% Large Language Models (LLMs) benefit from Retrieval-Augmented Generation (RAG) for accessing external knowledge, yet 
% Existing text-based RAG and graph-based RAG methods often fail under noisy and low-quality contexts, like conversational data. 
% Existing text-based RAG and graph-based RAG methods often fail under noisy and low-quality contexts, like conversational data. 
% We propose \emph{Structure-aware Retrieval-Augmented Generation} (\textbf{SA-RAG}), which introduces tables as an intermediate structured representation with quality-aware structure induction and generation. By providing a compact and normalized interface, SA-RAG effectively reduces noise while preserving essential information to improve performance.

We identify \textbf{structural quality under noisy conversational evidence} as a fundamental bottleneck of conversational RAG and propose  \textbf{SA-RAG} to address this challenge. SA-RAG effectively reduces retrieval noise while preserving essential information for downstream reasoning and generation. Extensive experiments demonstrate that improving \textbf{structural quality} consistently enhances retrieval robustness and response correctness.
\nocite{Ando2005,andrew2007scalable,rasooli-tetrault-2015}

% \section*{Limitations}
% A limitation of our method is the additional cost of inducing structured representations from raw conversational data. Compared with conventional RAG pipelines that operate directly on text, our approach requires a preprocessing stage to generate metadata and tables. This introduces extra computational overhead during data preparation. However, since this process is performed offline and only once per corpus, it does not impact the latency of online inference. We view this as a practical trade-off for improved robustness under noisy conversational data.
\section*{Limitations}
Our framework introduces an additional preprocessing stage to induce metadata and structured table representations from raw conversational data. Compared with conventional text-based RAG pipelines that directly operate on unstructured text, this step incurs extra offline computational cost during corpus preparation. SA-RAG currently relies on mostly static structure induction and generation, while online dialogue shifts require incremental refinement and structural-drift detection as discussed.
\bibliography{custom}

\appendix

\newpage

\section{Implement Details}
\label{app.implement}

We use \textbf{LLaMA-70B} (Llama-3.3-70b-instruct) and \textbf{Qwen-1.5B} (Qwen2-1.5B) as our base models. \textbf{GPT-4o} (gpt-4o) is employed as the \emph{LLM-as-a-judge}, with human verification conducted on a subset of examples to ensure evaluation reliability.
For LLaMA-70B, we adopt a training-free table generation approach, while Qwen-1.5B uses a training-based method to generate structured tables from dialogues. During online inference, the models retrieve relevant table rows and conversation snippets conditioned on the user query, and produce grounded answers.
Experiments are conducted on two datasets covering diverse conversational and retrieval-augmented generation (RAG) scenarios. Performance is evaluated using three metrics that jointly assess retrieval quality and answer generation: \textbf{Recall@3}, \textbf{Mean Reciprocal Rank@3 (MRR@3)}, and \textbf{Result Correctness}. 

\paragraph{Recall@3} For a query $q$ with a set of relevant items $R_q$, and a model returning the top-3 retrieved items $\text{top}_3(q)$, Recall@3 is defined as:

\[
\text{Recall@3}(q) = \frac{ \left| \text{top}_3(q) \cap R_q \right| }{ |R_q| }
\]

The overall Recall@3 across all queries $Q$ is the average:

\[
\text{Recall@3} = \frac{1}{|Q|} \sum_{q \in Q} \text{Recall@3}(q)
\]

\paragraph{MRR@3} For a query $q$ with at least one relevant item, let $\text{rank}_q(i)$ denote the rank of item $i$ in the top-3 retrieved list. The reciprocal rank for $q$ is:

\[
\text{RR@3}(q) = \frac{1}{\min\{ \text{rank}_q(i) \mid i \in \text{top}_3(q) \cap R_q \}}
\]

The Mean Reciprocal Rank@3 (MRR@3) is the average over all queries:

\[
\text{MRR@3} = \frac{1}{|Q|} \sum_{q \in Q} \text{RR@3}(q)
\]

\paragraph{Result Correctness} measures the factual accuracy and informativeness of the generated answers with respect to the retrieved evidence and the user query. It is scored using the \emph{LLM-as-a-judge} protocol, with manual verification on a subset of examples to ensure evaluation quality.

\subsection{Table Quality Evaluation}

To evaluate the quality of tables generated by our \textbf{SA-RAG} framework, we compute four metrics using concrete procedures:

\begin{enumerate}
    \item \textbf{Structural compliance of metadata}: 
    \begin{itemize}
        \item For each generated metadata, we check that all required information exist and are named correctly according to the structure requirement.

        \item A metadata receives a score equal to the fraction of columns that are present and correctly generated.
    \end{itemize}

    \item \textbf{Metadata effectiveness for downstream tasks}:
    \begin{itemize}
        \item For each column, we attempt to retrieve relevant information given sample user queries from the dialogue.
        \item If the column allows successful retrieval and provide useful information, the column is considered effective.
    \end{itemize}

    \item \textbf{Value-level constraint satisfaction}:
    \begin{itemize}
        \item We define constraints for each column, such as type constraints, value ranges, allowed categories, and format consistency.
        \item Each table cell is checked against its column constraints.
        \item The score for this metric is the fraction of cells that satisfy all constraints.
    \end{itemize}

    \item \textbf{Semantic correctness of table values}:
    \begin{itemize}
        \item Each table value is compared to the original dialogue to verify factual consistency.
        \item Incorrect or hallucinated values are flagged.
        \item The semantic correctness score is calculated as the proportion of table entries that are factually consistent with the dialogue context.
    \end{itemize}
\end{enumerate}

Finally, for each table, we average the four metric scores to obtain an overall table quality score. We perform this evaluation across all generated tables and compare the results of SA-RAG with two baselines: (1) predefined table generation and (2) direct table construction. This allows us to quantify improvements in structural validity, usability, and semantic fidelity of the generated tables. These metrics collectively assess both the structural and functional quality of the generated tables. Structural compliance ensures that the tables follow the required schema, which is critical for downstream tasks that rely on predictable formats. Metadata effectiveness measures whether the generated columns actually help answer user queries, reflecting the practical utility of the tables. Value-level constraint satisfaction guarantees that individual entries adhere to expected types and ranges, reducing errors and improving reliability. Finally, semantic correctness ensures that table contents are factually consistent with the source dialogue, preventing hallucinations and maintaining trustworthiness. Evaluating across these dimensions provides a comprehensive picture of table quality, highlighting improvements in usability, accuracy, and overall reliability achieved by our \textbf{SA-RAG} framework.
In this subsection, we will introduce our experiment setup, and please refer to Appendix \ref{app.implement} and \ref{app.data} for more information.

\paragraph{Application Scenario and Dataset Introduction}
We focus on information-seeking scenarios for conversational agents where users describe problems in free-form dialogue, and systems are required to organize, retrieve, and reason over historical data to generate accurate responses. Typical applications include technical support, customer service, and helpdesk systems, where conversations naturally involve structured elements such as issue types, causes, constraints, and solutions, but these structures are often implicit, incomplete, or noisy in the raw conversation.
We use two datasets, MSDialog and Customer Service Dialogue Dataset~\cite{InforSeek_User_Intent,InforSeek_Response_Ranking,InforSeek_User_Intent_Pred}.
The MSDialog dataset is a labeled public dialog corpus of interactions between information seekers and answer providers. The Customer Service Dialogue dataset is a private dataset with dialogues between customers and the service people in a highly professional domain. Customers raise issues, and service personnel provide detailed solutions.

\paragraph{Baselines} 
We evaluate our method against different competitive baselines spanning text-based, graph-based, IE-based and table-based retrieval-augmented generation (RAG) approaches.
Text-based RAG methods retrieve and condition solely on unstructured textual documents. We consider two variants: Text-freq, which retrieves documents based on term-frequency statistics and feeds the top-ranked texts to the LLM, and TextRAG-emb, which performs dense retrieval using text embeddings~\cite{lewis2021retrievalaugmentedgenerationknowledgeintensivenlp,10.5555/106765.106782}. 
% \RL{I think it may be necessary to describe the applicable scenarios for the method. In Figure 2, the example you provided also describes issue types. So is your experiment limited to tasks that facilitate extracting issue types/causes/answers?}
Graph-based RAG methods leverage a graph as the retrieval source to support the retrieval process. We include GraphRAG-KG, which retrieves entities directly from a knowledge graph, and GraphRAG-Rel, which explicitly retrieves and reasons over relation paths~\cite{wang2025knowledgegraphaugmentedlarge,edge2024local}.
Information extraction (IE) approaches directly extract structured knowledge from raw input text prior to retrieving useful information. We evaluate IE-rel, which extracts relational information using learned relation extraction models, and IE-rule, which relies on rule-based extraction pipelines to construct structured facts~\cite{yates-etal-2007-textrunner,stanovsky-etal-2018-supervised}.
Table-based generation methods operate on structured tabular representations. We consider TableRAG-pre, which retrieves relevant tables from predefined data and treats them as input to the LLM for generation, and TableRAG-dir, which directly generates table conditioned on conversational data~\cite{Herzig_2020,liu2022tapextablepretraininglearning}.

\subsection{Online Retrieval}

Given the structured table collection
$\mathcal{T} = \{T_1, T_2, \ldots\}$
and a user query $q$, we perform retrieval
and reasoning with the high-quality metadata and structured table , explicitly modeling the knowledge over structured tables and conversations before final generation.

We first perform dense retrieval over table rows.
Each table $M_i$ and the query $q$ are embedded into a shared vector space
via an encoder $\phi(\cdot)$.
The initial candidate set is defined as
$
\mathcal{T}_q^{(0)} =
\operatorname{TopK}_{T_i \in \mathcal{T}}
\ \text{sim}\!\left(\phi(q), \phi(T_i)\right),
$
where $\text{sim}(\cdot,\cdot)$ denotes cosine similarity.
We formalize information extraction as an iterative state transition process.
At step $t$, the information set is
$
S^{(t)} = \big(E^{(t)}, R^{(t)}, I^{(t)}\big),
$
where $E^{(t)}$ denotes the active values set,
$R^{(t)}$ the activated relations,
and $I^{(t)}$ the accumulated information.
The initial state $S^{(0)}$ is constructed from $\mathcal{M}_q^{(0)}$ and $q$.

% For a table $T_i \in \mathcal{T}_q^{(t)}$ with metadata
% $M^{\star} = \{m_1, \ldots, m_{|M|
% }\}$,
% we extract values associated with the user query and final answers.
For an value $e \in E^{(t)}$ and relation $r$,
the corresponding information is obtained by
$
v = \text{Tail}(e, r),
$
where $\text{Tail}(\cdot)$ retrieves the table entry.
Values are aggregated as
$
I^{(t+1)} = I^{(t)} \cup \{i\}.
$
To discover additional evidence, we expand the reasoning state by selecting new values at each step. A relation scoring function evaluates candidate relations conditioned on the query and the current state. The selected relation is added to the active set, and values connected through this relation are activated, forming the value set for the next iteration, i.e., $E^{(t+1)} = \{ e' \mid \exists e \in E^{(t)},\ \text{Tail}(e, \text{rel}^{(t)}) = e' \}$.
The information search process terminates when either a maximum depth is reached or sufficient evidence has been accumulated. 
Upon termination, we construct a structured evidence set
$
K_q = \{(e, r, i) \mid e \in E^{(t)}, r \in R^{(t)}, i \in I^{(t)}\},
$
explicitly representing value--relation--information triples
grounded in table structures.

The final response is generated by conditioning the LLM
on the query and the constructed structured knowledge:
\begin{equation}
\hat{a} = {Response Generator}(q, T_q,K_q).
\end{equation}
This formulation ensures that generation is grounded
in iteratively retrieved and verified structured evidence.
\subsection{Metadata Quality Objective.}
Given a set of normalized candidate metadata attributes
\[
\tilde{M}
=
\{\tilde{m}_{ij}\},
\]
we formulate metadata construction as a quality-aware optimization problem:

\[
L(M)
=
\sum_{i,j}
R(\tilde{m}_{ij}, D_i)
+
S(M),
\]

where $R(\cdot,\cdot)$ measures the usefulness of a metadata attribute with respect to the original conversational evidence, and $S(\cdot)$ evaluates the global structural consistency and normalization quality of the metadata schema.

\vspace{0.5em}

For each normalized candidate metadata attribute $\tilde{m}_{ij}$ generated by the LLM from conversation $D_i$, we define
\[
R(\tilde{m}_{ij}, C)
\]
as the LLM-estimated relevance and coverage score of $\tilde{m}_{ij}$ with respect to the useful information in the conversation corpus $C$.

Specifically, the LLM is prompted to judge whether the candidate metadata attribute:
\begin{enumerate}
    \item captures information that is explicitly or implicitly mentioned in the conversation;
    \item is useful for answering downstream user queries;
    \item is not redundant with existing metadata attributes;

\end{enumerate}

Thus, $R(\tilde{m}_{ij}, C)$ is not a differentiable learned function, but an LLM-generated quality score or categorical judgment used to guide metadata selection.

Given the current metadata set $M$, we define
\[
S(M)
\]
as the LLM-estimated structural normalization quality of the metadata schema.

The LLM evaluates whether the metadata set $M$ satisfies the following criteria:
\begin{enumerate}
    \item \textbf{Consistency}: semantically equivalent attributes are normalized into the same canonical form;
    \item \textbf{Non-redundancy}: duplicated or highly overlapping attributes are removed or merged;
    \item \textbf{Completeness}: the metadata schema covers the major types of useful information appearing in the conversations;

\end{enumerate}

Therefore, $S(M)$ represents an LLM-generated assessment of schema-level quality rather than a manually specified differentiable regularization term.

\paragraph{Metadata Update Operations.}
To maintain a compact and effective metadata schema, we define several update operations for each normalized metadata candidate $\tilde{m}_{ij}$:
% \[
% Q(\tilde{m}_{ij}, M^{(t)}, r^{(t)}_{ij}) =
% \begin{cases}
% \texttt{KEEP}, & \text{if semantically equivalent to existing metadata,} \\
% \texttt{EXPANDABLE}, & \text{if complementary to existing metadata,} \\
% \texttt{NEW}, & \text{if introducing a useful new attribute,} \\
% \texttt{OTHER OPERATIONS}, & \text{e.g., MERGE, UPDATE, DELETE.}
% \end{cases}
% \]
Here, \texttt{KEEP} preserves existing metadata (e.g., \textit{issue\_type} and \textit{problem\_category}), \texttt{EXPANDABLE} extends existing schema semantics (e.g., \textit{error\_code} $\rightarrow$ \textit{installation\_error\_code}), and \texttt{NEW} introduces previously unseen but useful metadata fields (e.g., \textit{operating\_system\_version}). Additional governance operations, such as \texttt{MERGE} and \texttt{DELETE}, maintain metadata quality and compactness.
\newpage
\onecolumn
\subsection{LLM Prompts for Table and Metadata Generation}

Below we provide the prompts used in our pipeline for structured table extraction, schema normalization, quality evaluation, and meta-data evolution. Each prompt is presented in a separate box for clarity.

\begin{tcolorbox}[colback=gray!5!white,colframe=black!75!black,title=Prompt 1: Extract Columns from Dialogue]
You are a data modeling assistant.

TASK: Given a customer support dialogue, infer a relational database table schema to solve the problem. Provide more than ten columns useful to classify and answer future issues, not redundant information.

INSTRUCTIONS:
\begin{itemize}
    \item Identify the main table that represents the dialogue.
    \item Generate a table schema using the following JSON structure ONLY:
\begin{verbatim}
[
  {
    "table_name": string,
    "Columns": [
      {
        "name": string,
        "type": one of ["int", 
        "string", 
        "float", "boolean", "date", 
        "datetime"],
        "semantic": string,
        "constraints": [string],
        "Note": "Important
        information"
      }
    ],
    "functional_denpendencies": 
    [string],
    "version": int
  }
]
\end{verbatim}
    \item Column names: snake\_case, concise, descriptive.
    \item Semantic: describe real-world meaning, prefer ontology-style naming.
    \item Constraints: use primary key, foreign key, not null, unique when applicable.
    \item Output valid JSON only, no explanations.
\end{itemize}

DIALOGUE: \{dialog\}
\end{tcolorbox}

\begin{tcolorbox}[colback=gray!5!white,colframe=black!75!black,title=Prompt 2: Normalize Schema]
You are a database schema normalization assistant.

TASK: Normalize the given raw database schema in ONE step.

RULES:
\begin{itemize}
    \item Column names: snake\_case, concise, descriptive.
    \item Data types: normalize to one of ["int", "string", "float", "boolean", "date", "datetime"].
    \item Semantic: provide clear standardized meaning, prefer ontology-style naming.
    \item Constraints: preserve valid constraints, normalize naming.
    \item Structure: preserve table\_name, version, functional dependencies; fix spelling issues.
    \item Output valid JSON only.
\end{itemize}

INPUT SCHEMA: \{raw\_schema\}
\end{tcolorbox}

\begin{tcolorbox}[colback=gray!5!white,colframe=black!75!black,title=Prompt 3: Quality Check]
You are a database schema quality evaluator.

TASK: Evaluate the quality of EACH COLUMN in the new schema based ONLY on its usefulness for answering the given problem.

COLUMN-LEVEL METRICS (0–1):
\begin{enumerate}
    \item relevance: How directly the column relates to the problem.
    \item answerability: Contribution to answering the problem.
    \item overall: Holistic judgment for this problem.
\end{enumerate}

Output: For each column, add field \texttt{quality\_score} with relevance, answerability, overall, and justification. Return valid JSON only.

CURRENT META DATA: \{current\_meta\_data\}

NEW SCHEMA TO EVALUATE: \{new\_schema\}

PROBLEM / QUERY: \{problem\}
\end{tcolorbox}

\begin{tcolorbox}[colback=gray!5!white,colframe=black!75!black,title=Prompt 4: Schema Governance and Merge]
You are a schema governance and evolution assistant.

TASK: Given current meta-data, a new schema with column-level quality scores, and problem context, decide how to evolve the meta-data to retain columns useful for identifying or solving the issue. Every column should be useful.

Steps (perform all in one pass):
\begin{enumerate}
    \item Decide if the new schema is acceptable.
    \item Detect semantic overlap with existing schemas.
    \item For each overlap, choose one operation: ADD, UPDATE, MERGE, KEEP, DELETE.
    \item Produce the FINAL updated meta-data (same format as current meta-data, without quality scores).
    \item Ensure no more than 20 columns.
\end{enumerate}

Decision rules:
\begin{itemize}
    \item Use column-level quality\_score.overall as primary signal.
    \item Prefer schemas with higher average column quality, better coverage, and clearer semantics.
    \item Delete only strictly worse redundant columns.
    \item Avoid introducing new tables.
\end{itemize}

Output: valid JSON ONLY, final meta-data list.

CURRENT META DATA: \{current\_meta\_data\}

NEW SCHEMA (WITH QUALITY SCORES): \{quality\_annotated\_schema\}

PROBLEM CONTEXT: \{problem\}
\end{tcolorbox}

\begin{tcolorbox}[colback=gray!5!white,colframe=black!75!black,title=Prompt 5: Generate Data Row from Dialogue]
You are a data extraction assistant.

TASK: Generate ONE structured data row from the given dialogue.

IMPORTANT RULES:
\begin{itemize}
    \item Meta-data is READ-ONLY context.
    \item Do NOT modify, normalize, rename, or reinterpret the schema.
    \item Use column names EXACTLY as provided.
    \item If a value cannot be inferred, output null.
    \item Follow declared data types strictly.
    \item Return VALID JSON ONLY.
    \item No explanations, no comments.
\end{itemize}

META-DATA: \{meta\_data\}

DIALOGUE: \{dialogue\}

OUTPUT FORMAT:
\begin{verbatim}
{
  "row": {
    "<column_name>": <value | null>
  }
}
\end{verbatim}
\end{tcolorbox}

\begin{tcolorbox}[colback=gray!5!white,colframe=black!75!black,title=Prompt 6: Judge Table Against Metadata]
You are a data validation assistant.

TASK: Determine whether the table satisfies the metadata requirements (e.g., all non-null columns are filled).

META-DATA: \{meta\_data\}

TABLE: \{dialogue\}

OUTPUT FORMAT:
\begin{itemize}
    \item Satisfy the metadata: yes or no.
    \item If no, explain which parts do not satisfy the metadata.
\end{itemize}
\end{tcolorbox}
\twocolumn

\newpage
\section{Dataset Introduction}
\label{app.data}
The MSDialog dataset is a labeled dialog corpus of question-answering (QA) interactions between information seekers and answer providers on the Microsoft Community forum~\cite{InforSeek_User_Intent,InforSeek_Response_Ranking,InforSeek_User_Intent_Pred}. It comprises over 2,000 multi-turn information-seeking conversations, totaling approximately 10,000 utterances, each annotated with user intent at the utterance level. Annotations were collected via crowdsourcing using Amazon Mechanical Turk.
MSDialog provides several versions, including the full collection (MSDialog-Complete) and a labeled subset (MSDialog-Intent).
For user intent annotation, they selected a subset of dialogs from MSDialog-Complete based on the following quality criteria: (1) conversations containing 3 to 10 turns, (2) involving 2 to 4 participants, (3) containing at least one community-verified correct answer, and (4) belonging to one of four major Microsoft product categories: Windows, Office, Bing, and Skype. This filtering resulted in approximately 2,400 dialogs for annotation.

We use the Customer Service Dialogue Dataset, a private multi-turn conversational corpus between customers and service personnel in a high-professional domain. It has 1000 conversations between customers and professional service people. Each conversation contains 1-5 turns between two participants. Each conversation records customers describing issues and service agents providing detailed solutions. While the dialogues often contain structured elements such as issue types, causes, constraints, and suggested resolutions, these structures are typically implicit, incomplete, or noisy in the raw conversation. Due to confidentiality requirements, we cannot disclose specific statistics such as the number of dialogues, turns, or users, but the dataset covers a wide range of real-world customer support scenarios, making it suitable for evaluating information retrieval, structured reasoning, and grounded response generation in conversational agents.

\subsection*{Preference Pair Construction}

For each conversation $d_i$, we construct a set of preference pairs $(M_i^{+}, M_i^{-})$ to enable preference-based supervision.

\paragraph{Positive Tables $M_i^{+}$} 
The positive table $M_i^{+}$ is obtained from human-annotated, high-quality structured representations. These tables are carefully curated to ensure:
\begin{itemize}
    \item \textbf{Semantic correctness}: all table entries accurately reflect the information in the dialogue.
    \item \textbf{Schema consistency}: column names, data types, and structural completeness conform to the predefined table schema.
\end{itemize}

\paragraph{Negative Tables $M_i^{-}$} 
Based on each $M_i^{+}$, we generate a corresponding negative table $M_i^{-}$ by applying controlled perturbations to the structured fields. The perturbations are designed to produce tables that are syntactically valid but semantically or structurally misaligned with the dialogue. The methods include:

\begin{enumerate}
    \item \textbf{Random field drop}: randomly set some fields to \texttt{null} to simulate missing values.
    \item \textbf{Dialogue-based hallucination}: replace some field values with randomly sampled words from the dialogue to produce realistic but incorrect entries.
    \item \textbf{Field swapping}: swap values across columns to create structural misalignment.
    \item \textbf{Dialogue-inconsistent fill}: fill fields with randomly selected sentences from the dialogue, teaching the model that direct copying of dialogue may be incorrect.
\end{enumerate}

\paragraph{Combination of perturbations} 
For diversity, one or multiple perturbations are randomly applied per table instance. This ensures that negative tables remain well-formed syntactically while violating semantic or structural alignment, which provides effective supervision signals for preference-based training.

\paragraph{Implementation Details} 
The negative table generation process is implemented in Python. For each dialogue row, we randomly select a perturbation mode from \texttt{["drop", "hallucinate", "swap", "dialogue", "combo"]}. The \texttt{combo} mode applies multiple perturbations sequentially to create more challenging negative examples. After generating all negative tables, we save them to a CSV file (\texttt{structure\_bad.csv}) for downstream use.

\begin{itemize}
    \item \texttt{drop\_random\_values(row, drop\_rate)}: randomly sets a fraction of fields to \texttt{null}.
    \item \texttt{random\_hallucination(row, noise\_rate)}: replaces selected fields with randomly sampled words from the dialogue.
    \item \texttt{swap\_fields(row, swap\_pairs)}: swaps values between randomly selected pairs of columns.
    \item \texttt{dialogue\_inconsistent\_fill(row)}: fills a random field with a randomly chosen sentence from the dialogue.
\end{itemize}

This process produces a diverse set of preference pairs $(M_i^{+}, M_i^{-})$, which are then used for training and evaluating the model's ability to distinguish high-quality structured tables from noisy alternatives.

% \section{Deploy information}
% Our structure-aware Retrieval-Augmented Generation (SA-RAG) framework is implemented as the core retrieval module of InsightBot, an industrial AIOps system designed for real-time online question answering over large-scale historical ticket repositories. InsightBot has been productized and integrated into NEC’s BluStellar solution portfolio~\footnote{NEC BluStellar. https://www.nec.com/en/global/necblustellar/index.html}, which provides end-to-end digital transformation and operational intelligence solutions for industry and enterprise users. In production, InsightBot is deployed as part of NEC’s AIOps infrastructure to monitor and respond to online operational queries for NEC cloud and telecommunications services, handling thousands of requests on a daily basis~\footnote{ NEC AIOps / System Orchestration. https://jpn.nec.com/systemorchestration/}. Within this system, structured RAG serves as the central retrieval engine, enabling accurate and robust reference selection from large-scale, noisy operational data by leveraging structured metadata, temporal and condition awareness, and quality-aware retrieval mechanisms. This real-world deployment demonstrates that structured RAG can operate reliably at scale in mission-critical environments, effectively bridging advanced RAG research with practical AIOps applications.
\section{Case Study}
\label{app:case_study}

\subsection*{Scenario Overview}
This case study illustrates how the \textbf{SA-RAG} framework processes a multi-turn customer support conversation from offline table construction to online retrieval and response generation. The conversation is anonymized and simplified to focus on the workflow and structured representation rather than user or device details.

\subsection*
{Offline Phase: Metadata and Table Generation}
In the offline stage, SA-RAG constructs structured metadata from historical conversations. For each conversation:

\begin{itemize}
    \item \textbf{Metadata Extraction}: The system identifies key attributes of the conversation, such as \texttt{issue\_title}, \texttt{issue\_description}, \texttt{operating\_system}, \texttt{symptoms}, and \texttt{agent\_response}.
    \item \textbf{Normalization}: Extracted fields are normalized according to a predefined schema: column names are in \texttt{snake\_case}, data types are standardized, and semantic meanings are clarified.
    \item \textbf{Quality Control and update}: Each column is evaluated for relevance, answerability, and overall quality with respect to the conversation context. Low-quality or inconsistent entries are flagged or corrected.

    \item \textbf{Final Table Storage}: After normalization and quality validation, the structured tables are stored in a database to be used for online retrieval.
\end{itemize}

At the end of this phase, each historical conversation is represented as a structured table row containing the most informative fields, ready to support downstream retrieval tasks.

\subsection*{Online Phase: Retrieval and Grounded Response}
When a new user query or ongoing conversation occurs:

\begin{itemize}
    \item \textbf{Query Understanding}: The system interprets the user’s question in the context of the ongoing conversation.
    \item \textbf{Table Retrieval}: SA-RAG retrieves relevant rows from the offline-generated tables that best match the current query, considering semantic similarity and relevance.
    \item \textbf{Grounded Answer Generation}: Using the retrieved table rows and the conversation context, the model generates responses that are factual and aligned with the structured metadata.
    \item \textbf{Dynamic Updates}: If a new schema is generated or existing metadata needs adjustment, SA-RAG evaluates the new schema using quality scores, decides whether to add, update, merge, or discard columns, and updates the meta-data for future queries.
\end{itemize}

\subsection*{Example Walkthrough}
Consider a conversation where a user reports a login failure:

\begin{enumerate}
    \item \textbf{Offline Table Generation}:
    \begin{itemize}
        \item \texttt{issue\_title}: Login failure
        \item \texttt{issue\_description}: Unable to log in, receiving an error about settings
        \item \texttt{operating\_system}: Mobile OS
        \item \texttt{symptoms}: Login error message appears
        \item \texttt{agent\_response}: Device no longer supported
        \item \texttt{issue\_resolution\_status}: Unresolved
        \item Other fields such as \texttt{resolution\_steps} 
         may be empty if not mentioned.
    \end{itemize}

    \item \textbf{Online Retrieval}:
    \begin{itemize}
        \item When a new query about a similar login issue arises, SA-RAG retrieves this row from the offline table.
        \item The system focuses on the most relevant columns (\texttt{issue\_title}, \texttt{issue\_description}, \texttt{symptoms}) to generate a grounded response.
    \end{itemize}

    \item \textbf{Grounded Response Generation}:
    \begin{itemize}
        \item The model produces a response explaining the issue, referencing the relevant table information, ensuring factual consistency with the conversation history.

    \end{itemize}
\end{enumerate}

This workflow demonstrates how SA-RAG bridges unstructured dialogues and structured reasoning, enabling accurate, context-aware, and explainable responses in conversational information-seeking scenarios.

\section{LLM Evaluation Agreement}
\label{agreement}
To validate the reliability of LLM-based evaluation, we conduct a cross-model consistency analysis across multiple state-of-the-art LLMs. Specifically, we compare the evaluation scores produced by GPT-4o, Claude 3.5 Sonnet, and Gemini 3 using Spearman correlation coefficients.
The results demonstrate strong agreement across different evaluators. In particular, GPT-4o achieves a Spearman correlation of 0.98 with Claude 3.5 Sonnet and 0.78 with Gemini 3. These consistently high correlations indicate that the evaluation results are stable across different LLM judges rather than being dependent on a specific model.
Overall, the findings support the validity and robustness of the LLM-as-a-judge paradigm for evaluating conversational retrieval and generation quality in our experiments.

\section{Further Generalization and Structure Robustness Analysis}
\subsection{Granularity Analysis}
\label{app:gu}
\begin{figure}[!t]
    \centering
    % 将 width 改为 0.5\textwidth，即原来的一半
    \includegraphics[width=0.5\textwidth, trim=0 0 0 0, clip]{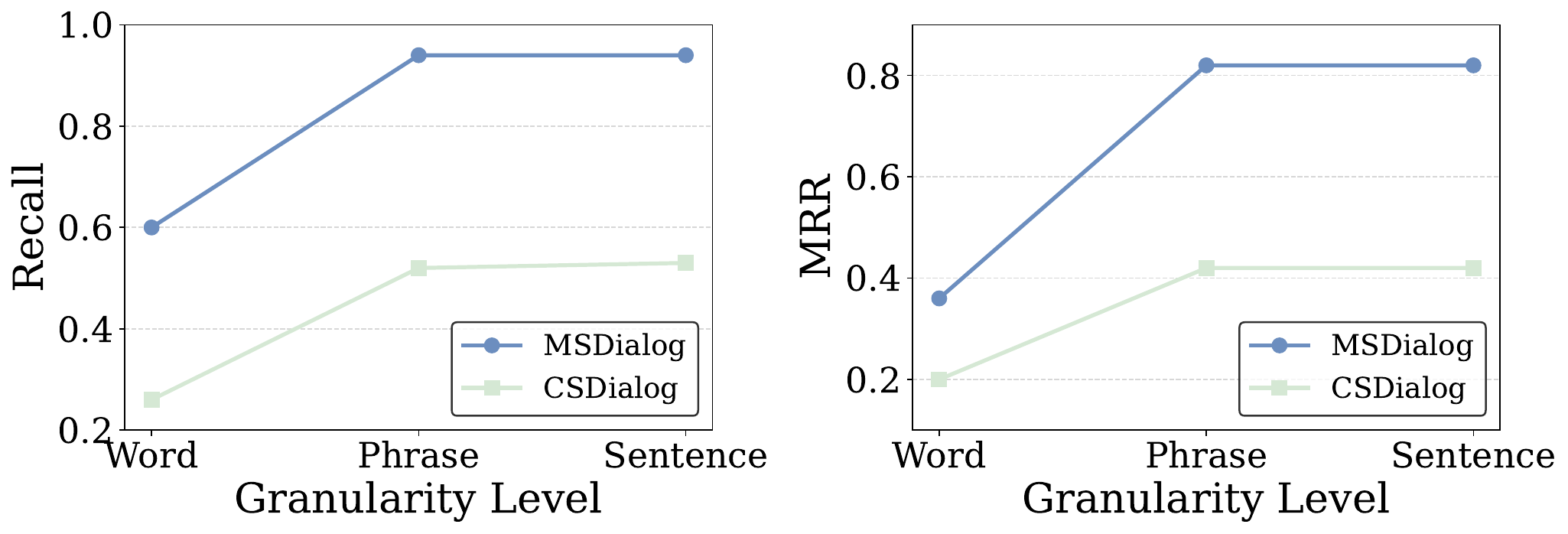}
\caption{Impact of structural granularity on retrieval performance.}
     \vspace{-1.5em} % 或 -1em 视情况调
    % \RL{The diagram is not visually appealing. A legend on the right could be used to show what each color represents.} \RL{The title can be improved to `Comparison of table quality metrics across table construction methods. Each subfigure reports one metric and higher scores indicate better performance.'}}
    \label{fig:gu}
\end{figure}
The granularity of table values plays an important role in structure-aware retrieval.
To this end, we conduct additional experiments by modeling values at different levels of granularity, including word-level, phrase-level, and sentence-level representations. Specifically, for word-level granularity, values are constructed from individual words, capturing fine-grained semantic signals but potentially introducing noise due to limited context. For phrase-level granularity, the model groups semantically coherent words, aiming to balance expressiveness and compactness. For sentence-level granularity, values correspond to full sentences, providing richer contextual information but at the cost of reduced specificity. To ensure a fair comparison, we keep the other setup the same as the original experiments, varying only the granularity of value construction. The results are summarized in Figure \ref{fig:gu}
Our findings show that:
Word-level representations lead to performance drops, due to insufficient contextual information.
While phrase-level and sentence-level representations achieve similarly strong performance, they better preserve semantic completeness.
This is because overly coarse granularity (word-level) breaks semantic completeness, while phrase/sentence-level representations better align with retrieval units.

\subsection{Additional Details on Out-of-Domain Generalization}
\label{ood}
We further evaluate SA-RAG on the Amazon Reviews dataset to study whether the proposed framework generalizes beyond conversational support scenarios. Compared with task-oriented dialogues, review data exhibits higher emotional density, sparse factual grounding, and more heterogeneous expressions, making structure induction and retrieval substantially more challenging.Amazon reviews are not structured as
problem-solution conversations, making this setting useful for testing whether
SA-RAG generalizes to noisy user-generated content.

Each retrieval unit is one review. A review is considered relevant if it
contains evidence supporting the answer to the query, such as product defect,
usage experience, sentiment reason, compatibility issue, or purchase-related issue
constraint. For table-based methods, each review is converted into a
metadata-aligned row, and retrieval is performed over generated table rows.
For text-based methods, retrieval is performed over raw review text.

Following the same retrieval and generation pipeline used in the main experiments, we construct a retrieval corpus consisting of 1,000 reviews and evaluate performance on 50 query instances. We compare SA-RAG with representative text-based, graph-based, and table-based retrieval baselines.

We evaluate different methods using five metrics. Recall measures the ability of the retriever to cover the correct answer, while MR evaluates the ranking quality by considering the position of the correct result in the retrieved list. Corr. denotes the final answer correctness. In addition, we introduce two metrics. ND (Noise Degradation) measures the performance drop after noise injection, where a lower value indicates stronger robustness against noise. Gain measures the relative improvement compared to the text-only baseline (\texttt{TextRAG-freq}), reflecting the effectiveness of incorporating structured information.

As shown in Table~\ref{tab:amazon_results}, \texttt{TextRAG-freq} achieves limited performance across Recall, MRR, and Corr., and exhibits a very high ND score of 0.98, indicating strong sensitivity to noise. By incorporating knowledge graph structures, \texttt{GraphRAG-KG} improves all effectiveness metrics and reduces ND to 0.79. \texttt{TableRAG-dir} further enhances retrieval and answer quality, achieving a Recall of 0.45 and a Corr. score of 0.41, while significantly reducing the degradation caused by noise injection. Among all methods, \texttt{SA-RAG} achieves the best overall performance, with Recall, MRR, and Corr. scores reaching 0.60, 0.52, and 0.57, respectively. At the same time, it reduces ND to 0.34, demonstrating substantially stronger robustness against noisy inputs. Moreover, \texttt{SA-RAG} obtains the highest Gain score of 0.55, indicating the largest improvement over the text-only baseline.

\section{Complexity and Scalability Analysis}
\label{app:c}
\paragraph{Computational Complexity.}
The computational complexity of metadata optimization is \(O(N)\), where \(N\) denotes the number of sampled conversations used during metadata induction. Notably, our framework does not require optimization over the entire corpus. Since metadata primarily captures invariant domain-level structural patterns, a representative subset of conversations is sufficient to induce a stable metadata schema that can subsequently be reused across the full dataset. This substantially reduces the practical computational overhead compared with corpus-wide optimization approaches.

In practice, we observe that the metadata architecture converges rapidly after processing a relatively small subset of conversations. Consequently, the cost of metadata induction becomes amortized over large-scale deployments, making the framework practical even for corpora containing millions of dialogue turns.

\paragraph{Parallelization.}
The proposed framework is naturally parallelizable and can be efficiently distributed across multiple computing nodes. Specifically, scalability is achieved through two complementary mechanisms:

\begin{enumerate}
    \item \textbf{Partition-Level Metadata Optimization:} 
    Conversation corpora are partitioned into independent subsets, and metadata optimization is performed concurrently on each partition.

    \item \textbf{Centralized Metadata Consolidation:} 
    The resulting local metadata structures are merged through a centralized normalization and validation stage, which resolves structural inconsistencies and aligns semantically equivalent attributes.
\end{enumerate}

Because metadata induction operates independently across partitions, the framework exhibits near-linear scalability with additional computational resources.

\paragraph{Incremental Processing.}
For newly arriving dialogue turns, the learned metadata is treated as a fixed schema. Instead of re-optimizing metadata globally, the system performs localized table population and attribute mapping only for the incoming data. This incremental mechanism eliminates the need for repeated corpus-wide reconstruction and significantly reduces online processing latency.

\paragraph{Schema Stability.}
Empirically, the metadata schema demonstrates strong stability once induced from a representative subset of the corpus. Most newly observed information can be accommodated within existing attributes and relations, thereby avoiding frequent schema restructuring. This property is particularly important for long-term deployment in continuously evolving conversational environments.

\paragraph{Triggered Refinement Mechanism.}
To maintain adaptability, lightweight metadata refinement is triggered only when the validation module detects substantial structural drift or persistent schema mismatch. Such updates are therefore sparse and localized, ensuring that adaptability does not compromise efficiency.

Overall, by decoupling offline structure induction from online retrieval and incremental updates, SA-RAG achieves both high scalability and strong adaptability for large-scale dynamic dialogue systems.

\section{AI Assistant Usage}
We used AI-based writing assistants to help polish the manuscript, including improving grammar, sentence fluency, and overall readability. All technical content, experimental design, analysis, and conclusions were developed and verified by the authors.

\end{document}